%% file: nips.tex
\documentclass{article}

\pdfoutput=1

\usepackage[preprint]{neurips_2025}

\usepackage[utf8]{inputenc} 
\usepackage[T1]{fontenc}    
\usepackage{hyperref}       
\usepackage{url}            
\usepackage{booktabs}       
\usepackage{amsfonts}       
\usepackage{amsmath}
\usepackage{graphicx}
\usepackage{multirow}
\usepackage{nicefrac}       
\usepackage{microtype}      
\usepackage{xcolor}  
\usepackage{wrapfig}
\newcommand{\vpara}[1]{\vspace{0.05in}\noindent\textbf{#1 }}

\newcommand{\Hide}[1]{} 

\usepackage{algorithm}

\usepackage{algpseudocode}

\title{Boosting Diffusion-Based Text Image Super-Resolution Model Towards Generalized Real-World Scenarios}

\author{
Chenglu Pan$^{1}$\footnotemark[3] \quad Xiaogang Xu$^{2}$\footnotemark[1] \footnotemark[2] \quad Ganggui Ding$^{1}$ \quad Yunke Zhang$^{1}$ \\
\quad \textbf{Wenbo Li}$^{2}$ \quad \textbf{Jiarong Xu}$^{3}$ \quad \textbf{Qingbiao Wu}$^{1}$\
\\
$^1$Zhejiang University \quad
$^2$The Chinese University of Hong Kong \\
$^3$Fudan University
\\
\small
\texttt{\{chenglupan,dingangui,yunkezhang,qbwu\}@zju.edu.cn} 
\\ \small \texttt{\{xiaogangxu00, fenglinglwb\}@gmail.com} \quad \small \texttt {jiarongxu@fudan.edu.cn}
}

\begin{document}

\maketitle
\renewcommand{\thefootnote}{\fnsymbol{footnote}}
\footnotetext[1]{Project Lead.} 
\footnotetext[2]{Corresponding Author.} 
\footnotetext[3]{This work is done during the internship of Chenglu Pan at Huawei.}

\input{sec/0_abstract}

\input{sec/1_intro}

\input{sec/2_related_work}

\input{sec/3_method}

\input{sec/4_exp}

\input{sec/5_conclusion}

\clearpage

\bibliographystyle{ieeenat_fullname}
\bibliography{nips}

\input{sec/6_appendix}
\end{document}

%% file: sec/0_abstract.tex
\begin{abstract}
    Restoring low-resolution text images presents a significant challenge, as it requires maintaining both the fidelity and stylistic realism of the text in restored images. Existing text image restoration methods often fall short in hard situations, as the traditional super-resolution models cannot guarantee clarity, while diffusion-based methods fail to maintain fidelity. 
    In this paper, we introduce a novel framework aimed at improving the generalization ability of diffusion models for text image super-resolution (SR), especially promoting fidelity. 
    First, we propose a progressive data sampling strategy that incorporates diverse image types at different stages of training, stabilizing the convergence and improving the generalization.
    For the network architecture, we leverage a pre-trained SR prior to provide robust spatial reasoning capabilities, enhancing the model's ability to preserve textual information. Additionally, we employ a cross-attention mechanism to better integrate textual priors. To further reduce errors in textual priors, we utilize confidence scores to dynamically adjust the importance of textual features during training. Extensive experiments on real-world datasets demonstrate that our approach not only produces text images with more realistic visual appearances but also improves the accuracy of text structure.
\end{abstract}

%% file: sec/1_intro.tex
\section{Introduction}
\begin{figure*}[t]
    \centering
     \includegraphics[width=0.95\textwidth]{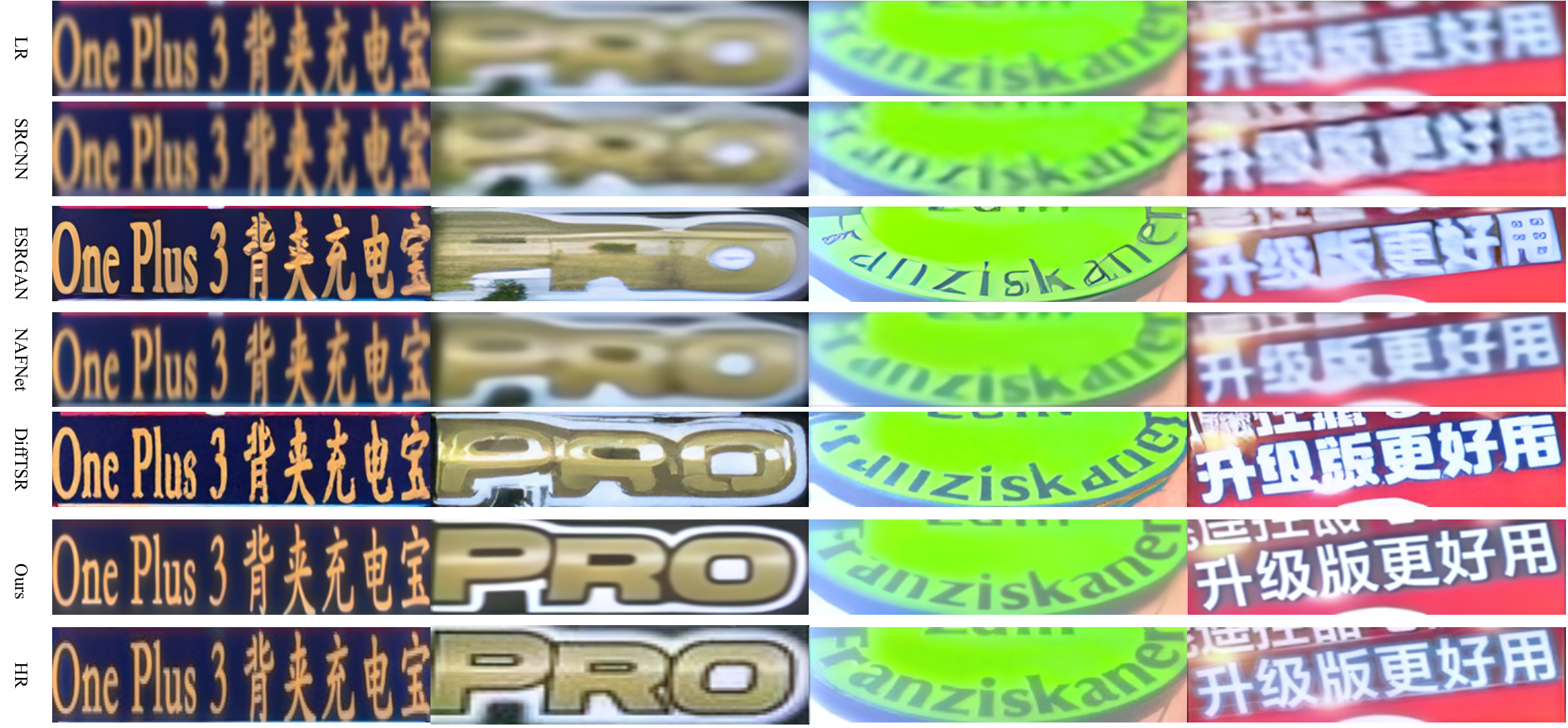}     
         \caption{Text image super-resolution results derived by different methods on real-world text images. Our model achieves superior accuracy with zero recognition errors while generating high-resolution text images, ensuring both precision and visual clarity.}
     \label{fig:tifo2}
\end{figure*}

Blind text image super-resolution (SR) is a task that aims to restore a high-resolution (HR) image with texts from a given low-resolution (LR) one. This problem is inherently more challenging than natural image SR, as it requires not only reconstructing the visual content but also strictly ensuring that the text’s fidelity and stylistic authenticity are preserved. Errors in the text structure can lead to semantic misinterpretations, which are unacceptable in the context of restored text images. This challenge is particularly evident when working with Chinese text, where even slight deviations in the stroke patterns of characters can lead to entirely different meanings, making the generated images look unnatural and unrealistic.

Existing SR methods for text images \cite{ma2022text, chen2021scene, ma2023benchmark, wang2020scene, ma2023text} leverage text priors to guide the restoration process, incorporating loss functions that enforce correct text structures. While these approaches perform well in many cases, they struggle with images that contain complex strokes or intricate fonts, resulting in inaccurate text feature restoration \cite{zhang2024diffusion}. Additionally, some methods improve architectures. E.g., MACRONet \cite{li2023learning} utilize a codebook to store character codes, which are helpful for generating structural details, and StyleGAN \cite{karras2020analyzing} is used to generate aesthetically pleasing text styles. However, these methods still fail when applied to LR images derived from unseen scenarios.

Recently, diffusion models (DM) \cite{ho2020denoising, sohl2015deep} have shown great promise in image restoration tasks, and researchers have started applying these models to text image SR \cite{zhang2024diffusion, singh2024dcdm}. These methods leverage the generative capabilities of DMs to restore images by modeling both the realistic image distribution and the underlying text prior. The combination of these two modalities significantly improves image quality compared to earlier methods. However, when dealing with real-world LR images, these methods still face significant challenges. Moreover, when OCR modules generate erroneous priors for LR images, the resulting super-resolved images may contain inaccurate or nonsensical text, which degrades the quality and clarity.

Furthermore, We observe that diffusion models trained solely on real data struggle to converge effectively, limiting SR performance improvement. This necessitates a dynamic sampling strategy combining real and synthetic data (e.g., from BSRGAN \cite{zhang2021designing} and Real-ESRGAN \cite{wang2021real}). These results highlight the need for SR approaches that can better adapt to varying real-world degradation levels.

To address these challenges, we propose a novel solution that enhances the model’s generalization ability through different aspects, including mixed training, new architecture, and confidence scores. 

The mixed training aims to stabilize the convergence and improve the generalization ability.
We begin by using degraded HR images as input to the model, establishing a strong foundation for text image restoration. Subsequently, we incorporate LR images from real-world scenarios into the training pipeline, enabling the model to adapt to a wide range of degradation types and improving its performance in more varied and realistic conditions. Finally, by progressively incorporating both degraded HR and real-world LR images into the training process, we enhance the model’s capacity to handle diverse real-world conditions, ultimately boosting its accuracy and generalization power.

Additionally, we argue that text image SR requires a more efficient network structure by using a SR-based diffusion rather than training the diffusion models from scratch. In this paper, we employ ResShift \cite{yue2024resshift} as an example to provide the SR prior. ResShift is a lightweight model that can be integrated with text modality inputs, enabling superior performance with fewer inference steps. This choice alleviates the challenges of fine-tuning large pre-trained models while maintaining high efficiency in both training and inference. We implement the interaction with text modality via the similar mechanism proposed in DiffTSR \cite{zhang2024diffusion}.

To further enhance the restoration quality, we propose the employment of confidence scores derived from an OCR module. By integrating confidence scores into the interaction with text modality, we can filter out low-confidence predictions and selectively refine only those areas where the model is uncertain. This mechanism helps reduce the occurrence of severe textual distortions by ensuring that the restoration process focuses on refining areas of the image that are more prone to error. As a result, this confidence-based approach helps to preserve text fidelity, ensuring that the final output is both accurate and visually coherent.

By focusing on improving generalization and adaptability, our method offers a robust solution to the challenges faced in text image SR. 
In summary, our contributions are as follows: 
\begin{itemize} 
    \item We introduce a novel mixed training strategy that progressively incorporates both degraded HR and real-world LR images to enhance the model’s generalization ability and improve its performance in diverse text image restoration tasks. 
    \item We reveal that a SR-based lightweight diffusion model is more suitable for text image SR, achieving better results and computational efficiency than the methods using large latent diffusion models that is trained from scratch. We further incorporate cross-attention from text priors and a confidence-guided mechanism, effectively enhancing text fidelity and structural consistency in super-resolution.
    \item Extensive experiments demonstrate that our model outperforms existing methods in the SR task for real-world text images, and achieves better generalization and restoration quality.
\end{itemize}

%% file: sec/2_related_work.tex
\vspace{-0.1in}
\section{Related Works}
\vspace{-0.1in}

\vpara{Blind Image Super-Resolution.}
Blind image super-resolution aims to recover high-quality images from low-resolution inputs that have undergone complex, unknown degradations in real-world scenarios. Recent advancements in this field have tackled the problem from two primary perspectives: degradation estimation \cite{bell2019blind,gu2019blind,huang2020unfolding,maeda2020unpaired,liang2022efficient,wang2021unsupervised} and real-world data synthesis \cite{cai2019toward,fritsche2019frequency,ji2020real,nazeri2019edgeconnect,li2022face,wang2021real,wei2020component,zhang2021designing}. The former focuses on estimating degradation models from LR images in an unsupervised manner \cite{wang2021unsupervised}, followed by the application of non-blind super-resolution methods. The latter, on the other hand, constructs LR-HR training pairs through a sophisticated degradation process that mimics real-world image deterioration. However, we observe that when it comes to text images, where characters exhibit specific and semantic text styles, relying solely on predefined degradation models is insufficient for achieving optimal restoration performance. Thus, using more advanced models like diffusion is required.

\vpara{Diffusion Models.}
Inspired by the non-equilibrium thermodynamics theory \cite{jarzynski1997equilibrium}, diffusion model \cite{sohl2015deep,ho2020denoising} is proposed to model complex distributions and has achieve impressive performance in various generative modeling tasks, including image/video generation \cite{ho2022video,saharia2022photorealistic,wu2022diffusion,ye2022first,zhang2024hive} and audio synthesis \cite{kong2020diffwave}.
Diffusion model aims to build the conditional distribution through a Markov Chain.\Hide{ where a sequence of small Gaussian noise is gradually added to the latent vector $z$ for $T$ steps, and the sizes of noise are controlled by a variance schedule $\{\beta_t\}_{t=1}^T$. Suppose that the latent data distribution is $q(z)$, the forward process is:
\begin{equation}
    q(z_t|z_{t-1})=\mathcal{N}(z_t;\sqrt{1-\beta_t}z_{t-1},\beta_t\textbf{I}).
\end{equation}
By using the reparameterization tricks, $z_t$ at any timestep $t$ can be sampled through
\begin{equation}
    q(z_t|z_0)=\mathcal{N}(z_t;\sqrt{\bar{\alpha_t}z_0,(1-\bar{\alpha_t}\textbf{I})}),
\end{equation}
where $\alpha_t=1-\beta_t$ and $\bar{\alpha_t}=\Pi_{k=1}^t\alpha_k$. Since $\beta_t$ is small enough, the distribution of reverse $q(z_{t-1}|z_t)$ is also Gaussian. Therefore the reverse step can also be modeled through a neural network $\theta$, denoted as $p_{\theta}(z_{t-1}|z_t)$:
\begin{align}
    p_{\theta}(z_{t-1}|z_t)=\mathcal{N}(z_{t-1}; \mu_{\theta}(z_t, t),\tilde{\beta}_t\textbf{I}),\\
    \mu_{\theta}(z_t,t)=\frac{1}{\sqrt{\alpha_t}}(z_t-\frac{1-\alpha_t}{\sqrt{1-\bar{\alpha_t}}}\epsilon_{\theta}(z_t, t)),
\end{align}
where $\tilde{\beta}_t=\frac{1-\bar{\alpha}_{t-1}}{1-\bar{\alpha}_t}\beta_t$, $\epsilon_{\theta}$ is the noise predicted by neural network, and the mean $\mu_{\theta}$ is derived through Bayes rules.
}
Benefiting from the strong power of data distribution modeling of diffusion models, plenty of works \cite{kawar2022denoising,wu2024one} also achieve impressive performance in image SR via the diffusion prior. However, they still have some failing cases.

\vpara{Text Image Super-Resolution.}
Text image super-resolution aims to enhance the fine details and clarity of text images, with a particular emphasis on improving text readability and recognition accuracy. Recent advances have largely focused on integrating text-specific priors and character structure priors to enhance the performance of text image SR. Specifically, prior works have leveraged text-related information from three primary perspectives: text-aware loss functions \cite{chen2021scene, wang2019textsr, wang2020scene}, text recognition priors \cite{ma2022text, xu2017learning, zhao2022c3}, and text structure priors \cite{li2023learning}. These approaches have demonstrated that text priors are essential for improving the structural integrity of text in the super-resolution process.
Despite the effectiveness of these priors, most existing methods fail to fully exploit the available text prior information, often struggling with challenges such as diverse text styles, severe degradation, or intricate strokes. Notably, diffusion-based models have recently been applied to text image SR \cite{singh2024dcdm, zhang2024diffusion}. 
In this paper, we propose that a SR-based diffusion prior together with a confident textural prior can achieve better performance and efficiency.

%% file: sec/3_method.tex
\section{Methods}

\begin{figure*}[t]
    \centering
     \includegraphics[width=0.9\linewidth]{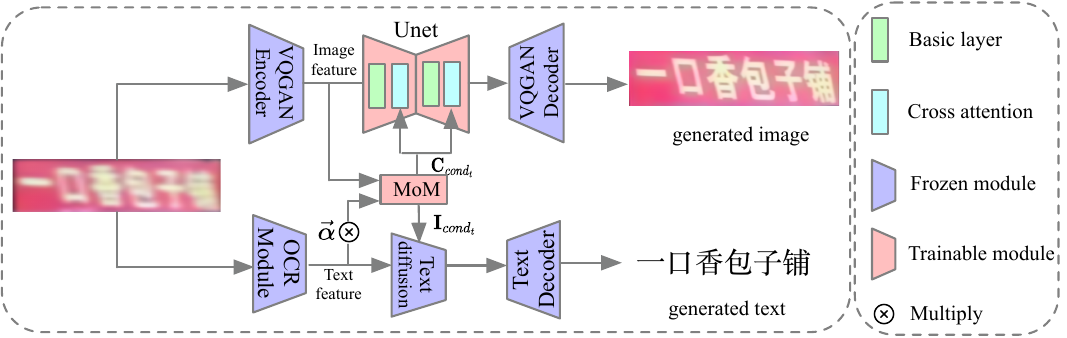}
         \caption{Overview of our framework. Our  model uses the confidence score $\vec{\alpha}$ to reduce the impact of false generation. $\mathbf{C}_{cond_t}$ generated by MoM from \cite{zhang2024diffusion} is fused with image feature via the cross attention module in the Unet module, which follows the basic layer in the pre-trained SR network.} 
     \label{fig:frame}
     \vspace{-0.2in}
  \end{figure*}

\subsection{Overview}
In this paper, we propose a method to estimate the HR image $x_0$ from the given LR image $y$. Our method is inspired by DiffTSR \cite{zhang2024diffusion} while has some distinct improvements. 
\Hide{
In our framework, an OCR module and a text diffusion model are also employed to recover the text information from the LR image, and a Mixture of multi-modality module (MoM) is used to facilitate the exchange of information between the image and text features.  
}
This section offers a comprehensive description of our approach. In Section \ref{subsec:data}, we propose a novel progressive data sampling strategy. Following that, in Section \ref{subsec:model}, we provide an in-depth explanation of our framework, with an illustrative overview presented in Figure \ref{fig:frame}. Finally, in Section \ref{subsec:loss}, we introduce the loss function that is used to optimize the model.

\subsection{Data Sampling Strategy During Training} \label{subsec:data}
\begin{wrapfigure}{h}{0.6\textwidth}
    \vspace{-1em}
    \centering
    \includegraphics[width=\linewidth]{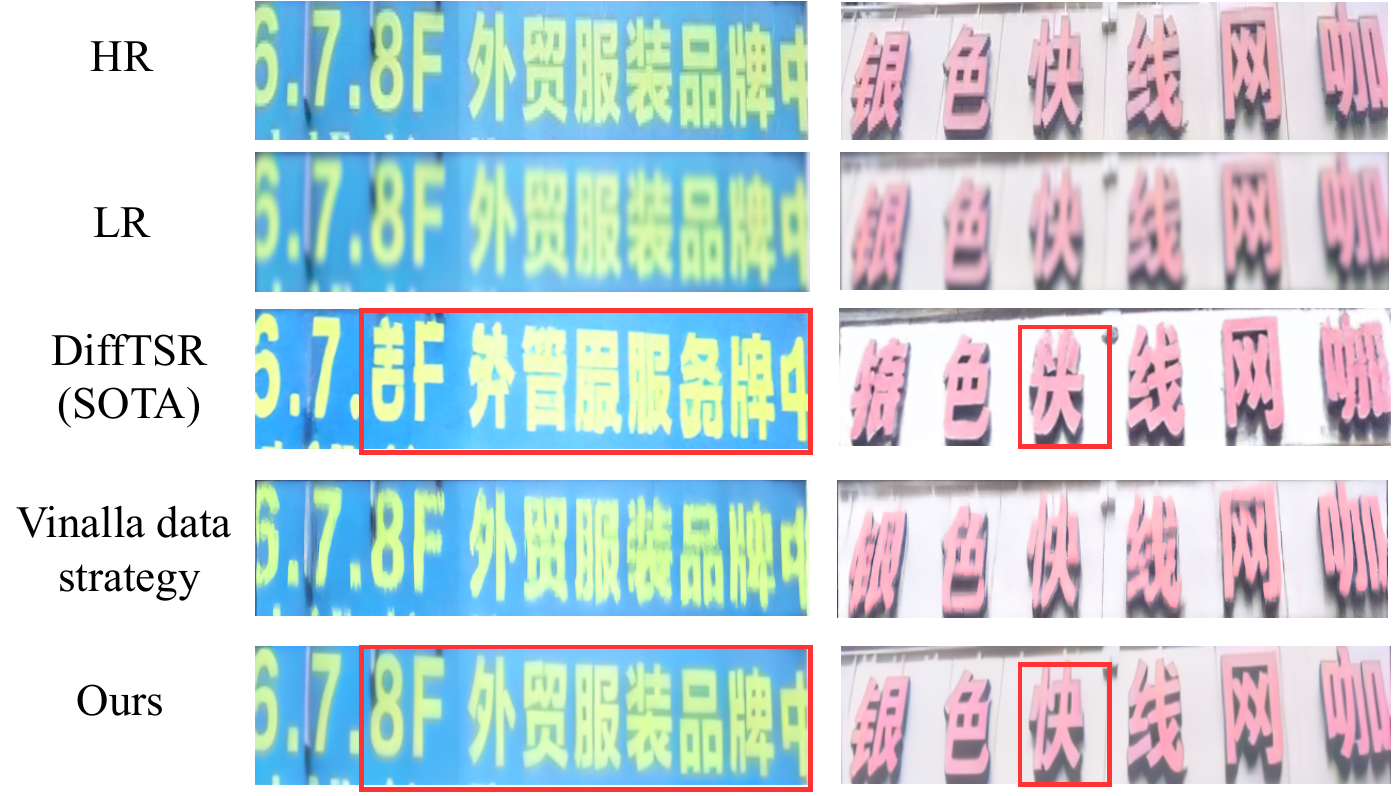}
        \caption{Results when training with only the real-world LR images (vanilla data strategy) and with our data strategy.}
    \label{fig:mot1}
\end{wrapfigure}

\noindent\textbf{Our Observations.} In this work, we address a key distinction from previous approaches \cite{zhang2024diffusion, singh2024dcdm, li2023learning} by focusing on improving the performance towards LR images derived from real-world scenarios. 
However, directly training the diffusion model with real-world LR images does not yield satisfactory results. As illustrated in Figure \ref{fig:mot1}, when the SR model is trained solely on real-world LR images, the generated outputs show minimal improvement over the input LR image, indicating that the model fails to converge. Moreover, when using degraded HR images as LR inputs, as done in methods like DiffTSR \cite{zhang2024diffusion}, the results are even more problematic—producing false text and suboptimal performance. This highlights the inherent limitations of relying on the sole source of LR images during the training.

\noindent\textbf{Overview.} In contrast to prior approaches, which predominantly rely on either real-world LR images or degraded HR images, our approach emphasizes a progressive training strategy designed to adapt to the complexities of real-world data. This strategy aims to gradually introduce varying levels of degradation into the training process, allowing the model to better learn from diverse scenarios. By leveraging both degraded HR images and real-world LR images, we enable the model to capture the subtle variations in image quality and text structure that are typical in practical applications.
Denoted the degraded HR images as $\hat{x_0}$, real-world LR images as $y$, and HR images as $x_0$.

\noindent\textbf{Our Implementation.} In our framework, we first conduct the training with the pair of $\hat{x_0}$ and $x_0$, initially building the model's core SR capabilities. 
After a few steps of training, we add the LR image $y$ into training by mixing $y$ and $\hat{x_0}$ when constructing the HR-LR pairs, which helps the model to improve the ability of in-domain adaptation. Specifically, during training, given the HR image $x_0$, the probabilities of forming an LR-HR pair using $y$ and $\hat{x_0}$ are equal.

Furthermore, after a sufficient number of steps with this configuration, we further extend the training process by introducing the degraded LR image, denoted as $\hat{y}$. This additional step occurs after learning from $y$ and $\hat{x_0}$, allowing it to gradually adapt to more degradation patterns. By incorporating $\hat{y}$, the model is exposed to increasingly complex degradation scenarios, improving its robustness in handling diverse LR inputs.

\noindent\textbf{Explanation.} This staged approach—starting with degraded HR images, followed by the LR image $y$, and then gradually introducing the degraded LR image $\hat{y}$—ensures a smooth and gradual progression in the model’s learning process. The model first learns to recover high-resolution details from relatively clean HR images, which serve as a strong prior for the restoration task. As the training progresses, the model is then exposed to the real-world LR image $y$, which introduces variability in the data distribution. By incorporating $y$, the model starts to develop the capacity for in-domain adaptation, learning to deal with the inherent degradation present in practical scenarios. The introduction of $\hat{y}$ further enhances the model’s ability to generalize to more out-of-domain and complex degradation patterns. This staged training procedure allows the model to not only preserve the high-frequency details from the HR images but also adapt effectively to the challenging degradation conditions typical of real-world LR inputs.
This is clearly demonstrated in the final row of Figure \ref{fig:mot1}, where the restored images show remarkable enhancement over the baseline, highlighting the effectiveness of our staged approach.

\subsection{Network Architecture} \label{subsec:model}
\Hide{Since DiffTSR \cite{zhang2024diffusion} leverages a diffusion model that is trained from scratch as the backbone U-Net, it becomes difficult to handle the LR images that is from the real-world scenarios.} We think, in the context of text image SR, leveraging a pre-trained SR model is more essential for restoring the fine details in LR images, while not requiring extensive generative capabilities. 
A model like ResShift \cite{yue2024resshift}, which primarily incorporates self-attention mechanisms through the Swin Transformer architecture, offers robust spatial reasoning capabilities for enhancing image quality. However, it lacks the necessary integration of external textual information, which is crucial for tasks that involve text image restoration.

\noindent\textbf{Diffusion Procedure.} In this work, we aim to improve text image SR by integrating both SR priors and text-specific priors, addressing the limitations of relying solely on pre-trained text-to-image diffusion priors.
While ResShift \cite{yue2024resshift} is an effective SR prior, it is inherently limited in its ability to get textural priors.
To overcome this, we introduce a cross-attention layer after each Swin Transformer block, enabling the model to integrate text information directly into the U-Net architecture. This cross-attention mechanism allows the model to not only perform high-quality SR but also perceive the semantic integrity of the text within the images. By doing so, the model gains the ability to restore text images with accurate and coherent textual features, addressing the unique challenges of text image restoration.

The process is carried out in the latent space, where the image $x_0$ is first mapped into a latent representation $z_0$ using VQGAN \cite{esser2021taming}. The forward process is defined as follows:
\begin{equation} \label{eq:1}
    q(z_t|z_{t-1},y)=\mathcal{N}(z_t;z_{t-1}+\alpha_t e_0,\kappa^2\alpha_t\textbf{I}),
\end{equation}
where $e_0 = y - x_0$ represents the residual between the LR and HR images, and $\{ \eta_t \}_{t=1}^T$ is the shifting sequence that monotonically increases with timestep $t$, with $\eta_1 \to 0$ and $\eta_T \to 1$. The hyper-parameter $\kappa$ controls the noise variance. 
The reverse distribution can be derived explicitly as \cite{yue2024resshift}:
\begin{equation} \label{eq:reverse}
    \footnotesize{
    q(z_{t-1}|z_t, y)=\mathcal{N}\left(z_{t-1}|\frac{\eta_{t-1}}{\eta_t}z_t+\frac{\alpha_t}{\eta_t}f_{\theta}(z_t, y, t),\kappa^2\frac{\eta_{t-1}}{\eta_t}\alpha_t\textbf{I}\right),}
\end{equation}
where $f_{\theta}$ is the deep neural network aiming to predict $z_0$.
This dual approach—combining ResShift for SR with cross-attention layers for text integration—offers a robust solution to the problem of text image super-resolution. By leveraging both SR priors and text priors, the model can handle complex text images that feature diverse fonts, structures, and degradation levels. Moreover, this method ensures that fine-grained text details and semantic coherence are maintained during the restoration process.

\noindent\textbf{Confidence Strategy.}
Another challenge occurs when the OCR module generates incorrect predictions for the LR image, as this may impede the image diffusion module's ability to correct the error, leading to the generation of incorrect characters with entirely different semantic meanings. Therefore, before training the model, we utilize a TransOCR \cite{chen2021scene} to get the confidence score of the output characters. Specifically, for each character $i$ in the image, the TranOCR module outputs the prediction $\hat{i}$ and the confidence score $\alpha_i$. $\alpha_i$ could be denoted as the predicted probability of the character $i$. Then the confidence scores of the text image could be formulated as
\begin{equation}
    \vec{\alpha}=[\alpha_1, \alpha_2,\cdots,\alpha_m],
\end{equation}
where $m$ is the length of the text. Then when we convert the text feature to MoM module in DiffTSR \cite{zhang2024diffusion}, we could multiply the text feature with the confidence scores of the text image, \emph{i.e.},
\begin{equation}
    [\textbf{I}_{cond_t}, \textbf{C}_{cond_t}]=\text{MoM}([z_y, z_t], \vec{\alpha}\cdot c_t, t),
\end{equation}
where $z_y$ is the latent representation of LR image $y$ mapped by VQGAN.
The method outlined above leverages the confidence scores from the TransOCR \cite{chen2021scene} module to adjust the importance of each character during training. By weighting the text features according to their confidence scores, the model can focus more on reliable characters while reducing the impact of erroneous ones.

\subsection{Loss Function} \label{subsec:loss}
To optimize the performance of our model, we combine three key loss functions: L1 loss, LPIPS loss, and cross-entropy loss for text recognition. Each of these components is designed to address different aspects of the text image super-resolution task, ensuring both high-quality image restoration and accurate text recovery.

First, the L1 loss is employed to minimize the pixel-wise differences between the generated super-resolved image $\tilde{x_0}$ and the ground truth HR image $x_0$. This term promotes fine-grained restoration by encouraging the model to reduce the absolute pixel differences.
Then we incorporate LPIPS \cite{zhang2018unreasonable} loss which enhances the perceptual similarity between the generated and ground truth images, denoted as $\mathcal{L}_{LPIPS}$. 
Unlike L1 loss, which operates at the pixel level, LPIPS captures higher-level semantic differences, improving the visual quality of the output.
Finally, the cross-entropy loss $\mathcal{L}_{CE}$ is introduced to ensure that the text in the generated image is semantically consistent with the ground truth text. This term is computed over the predicted characters and the true characters in the image.

The total loss function is the weighted sum of them:
\begin{equation}
    \mathcal{L}=\lambda_1\mathcal{L}_{L1}+\lambda_2\mathcal{L}_{LPIPS}+\lambda_3\mathcal{L}_{CE},
\end{equation}
where $\lambda_1, \lambda_2, \lambda_3$ are the respective hyperparameters controlling the importance of each term.

%% file: sec/4_exp.tex
\section{Experiments}
\subsection{Experimental Setup}
\vpara{Datasets}
In this work, we mainly focus on the SR of text characters in real-world scenarios as the exploration for complex characters, including Chinese characters. Following \cite{zhang2024diffusion}, we preprocess the image datasets with the following steps: (1) remove the images with a resolution smaller than 64 pixels; (2) only retain the images with a length of texts not larger than 24; (3) only retain the images with a width-to-height greater than 2; (4) resize the image to $128 \times 512$. Then, we have over 400K HR text images from the real-world scenario, and each of these images has a paired LR image. The degradation to generate $\hat{x_0}$ and $\hat{y}$ is proposed in BSRGAN \cite{zhang2021designing}. We divide the dataset into training and testing sets, with 80\% of the instances divided in the training set and the remaining instances divided in the testing set. We also include the CTR-TSR-Test dataset from \cite{yu2021benchmarking} for evaluation.

\vpara{Evaluation Metrics and Baselines.} To assess the performance of our proposed model, we compare it with several state-of-the-art image SR methods, including general SR approaches (\emph{e.g.,} ESRGAN \cite{wang2021real}, NAFNet \cite{chen2022simple}) and recent text image SR techniques (\emph{e.g.,} SRCNN \cite{dong2015image}, TSRN \cite{wang2020scene}, TSBRN \cite{chen2021scene}, MACRONet \cite{li2023learning}, DiffTSR \cite{zhang2024diffusion}). {When evaluating on the CTR-TSR-Test dataset, we only train our model on its corresponding training set. When testing on our collected dataset, for the baselines without training code, we adopt their released models}; for these with training codes, we finetune them with our collected dataset for a fair comparison.

We utilize five evaluation metrics to quantify the effectiveness of these methods. First, we compute the peak signal-to-noise ratio (PSNR) and the learned perceptual image patch similarity (LPIPS) \cite{zhang2018unreasonable}. To further assess the realism of the generated images, we report the Frechet Inception Distance (FID) \cite{heusel2017gans}. For a more specific evaluation of the text fidelity in restored text images, we incorporate word accuracy (ACC) and normalized edit distance (NED) \cite{ma2023benchmark}. Notably, we use the pre-trained TransOCR \cite{chen2021scene, yu2021benchmarking} as the text recognition model to compute ACC and NED.

\vpara{Implementation Details.} Our method is implemented using PyTorch. All experiments are conducted on a multi-GPU setup with four NVIDIA A100 GPUs, each equipped with 80GB of memory. We set the batch size to 64 and use the Adam optimizer \cite{kingma2014adam} with a learning rate of $1 \times 10^{-5}$. In the loss function, the hyperparameters $\lambda_1$, $\lambda_2$, and $\lambda_3$ are empirically set to 1, 1, and 0.02, respectively.

\begin{table}[t]
    \centering
    \caption{Quantitative comparison for the synthetic dataset on CTR-TSR-Test \cite{yu2021benchmarking} with different methods.}
    \resizebox{0.9\columnwidth}{!}
    {
    \begin{tabular}{l|ccccc|ccccc}
    \hline
    \multicolumn{1}{c|}{\multirow{2}{*}{method}} & \multicolumn{5}{c|}{$\times$ 2}        & \multicolumn{5}{c}{$\times$ 4}         \\ \cline{2-11} 
    \multicolumn{1}{c|}{}                        & PSNR $\uparrow$ & LPIPS $\downarrow$ & FID $\downarrow$ & ACC $\uparrow$ & NED $\uparrow$ & PSNR $\uparrow$ & LPIPS $\downarrow$ & FID $\downarrow$ & ACC $\uparrow$ & NED $\uparrow$ \\ \hline
    ESRGAN                                       & 24.75 &  0.191  & 9.308  & 0.8112 & 0.8239 & 20.90  & 0.310   &  21.86   &  0.6179   &  0.6272   \\
    SRCNN  &  23.73 & 0.338 &    54.47 & 0.7856 & 0.7991 & 20.74 & 0.501 & 116.5  & 0.6031 & 0.6160                                         \\
    NAFNet                                   & 25.04  &  0.286     & 37.42    &  0.8083   &  0.8212   &  21.82  &   0.447    & 87.93    & 0.6451    &  0.6573   \\
    TSRN                                   &  20.86    &  0.392     &  70.75   &  0.7805   & 0.7937    &    19.41  &   0.535    &   137.3  &  0.6149   &  0.6267   \\
    TBSRN                             &   24.43   &  0.282     & 57.61    &   0.8018  &  0.8156   &  21.56   &  0.442     &  132.6  &  0.6360   &  0.6486  \\
    MACRONet                       &  20.77    &   0.374    & 94.60    &  0.6934   &  0.7068   &   19.33   &   0.436    & 108.5    & 0.5123    &  0.5241   \\
    DiffTSR                            &  25.08    &  0.156     &  5.906   &  0.8594   &  0.8718   &   21.85   &  0.231     &  8.482   &  0.8350   &  0.8471   \\
    Ours                                            & \textbf{25.34} &\textbf{0.138}  &  \textbf{3.255}   &  \textbf{0.8779}   &  \textbf{0.8913}   &  \textbf{23.50}    &  \textbf{0.128}     &   \textbf{5.583}  &  \textbf{0.8511}   &   \textbf{0.8723}  \\
    \hline
    \end{tabular}
    }
    
    \label{tab:res1}
    \vspace{-0.1in}
\end{table}

\subsection{Quantitative Comparison}
We present a quantitative comparison on the synthetic CTR-TSR-Test dataset. As shown in Table \ref{tab:res1}, our method consistently outperforms all baseline approaches across all evaluation metrics. Specifically, it achieves the highest PSNR, demonstrating its ability to accurately reconstruct text images. Even under significant degradation, such as a $\times 4$ downsampling factor, our method maintains competitive performance. This can be attributed to the effective incorporation of a learned generative structure prior for each character, which enables robust recovery of text features even from severely degraded inputs.
Furthermore, our method excels in both LPIPS and FID, outperforming all baselines. 
The improved LPIPS suggests that our model preserves more fine-grained structural details, while the lower FID indicates that the generated images are more realistic when compared to real-world examples.
In terms of text fidelity, our method also surpasses DiffTSR in both ACC and NED, reflecting its superior ability to maintain the accuracy and structure of the restored text. The inclusion of the confidence score mechanism plays a crucial role, allowing the model to better preserve the integrity of the reliable textual priors during the generation process.

\begin{figure*}[h]
    \centering
     \includegraphics[width=0.95\linewidth]{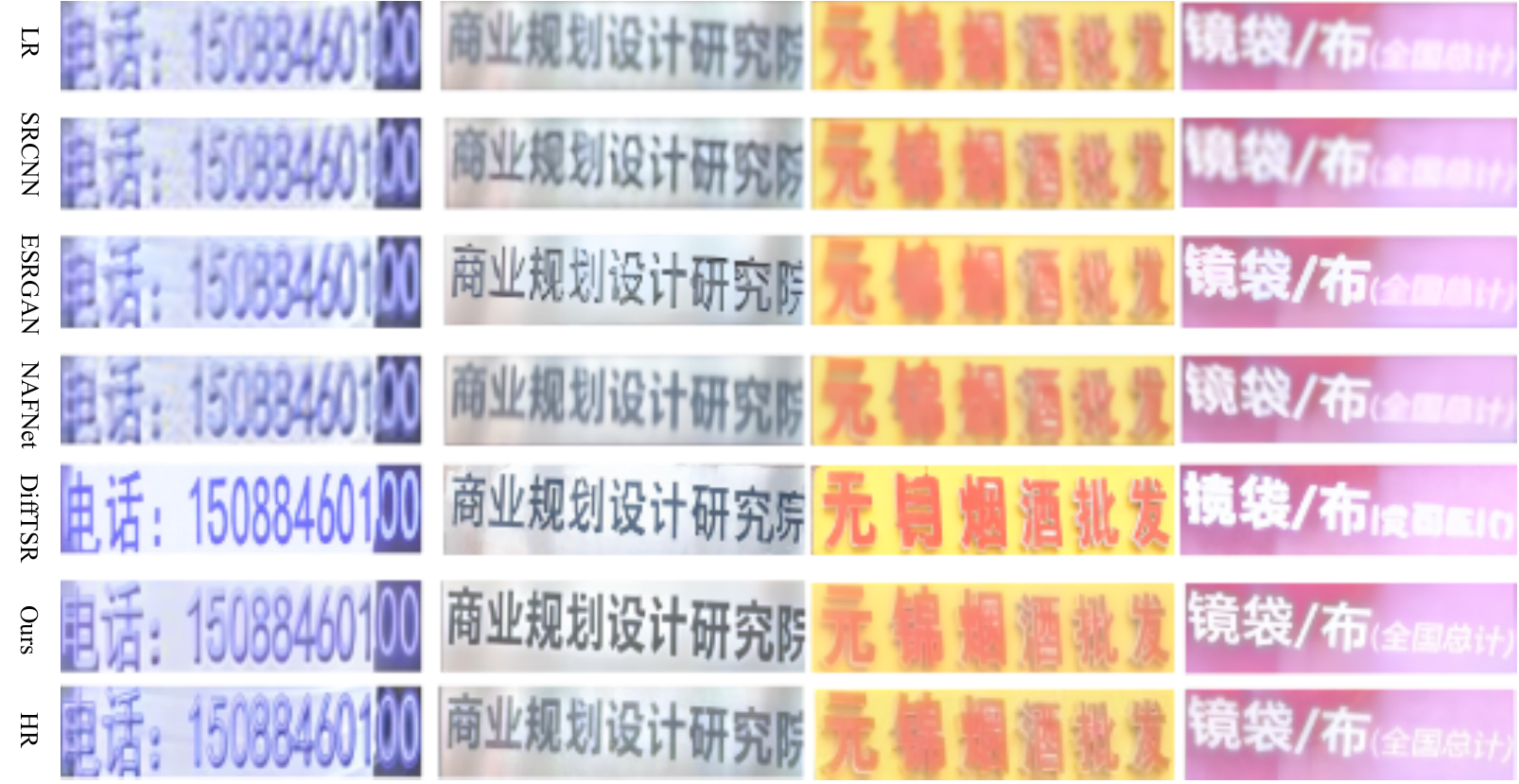}
         \caption{Qualitative comparison on the real-world dataset with different methods including SRCNN \cite{dong2015image}, NAFNet \cite{chen2022simple}, ESRGAN \cite{wang2021real}, the existing SOTA method DiffTSR \cite{zhang2024diffusion} and our method for $\times 4$ super-resolution. }
     \label{fig:tifo}
\end{figure*}

\subsection{Qualitative Comparison}
The qualitative results on the real-world dataset are shown in Figure \ref{fig:tifo}. Our method outperforms all other SR and text super-resolution techniques, including SRCNN \cite{dong2015image}, NAFNet \cite{chen2022simple}, ESRGAN \cite{wang2021real}, and the SOTA method DiffTSR \cite{zhang2024diffusion}, in generating high-quality text images. While ESRGAN \cite{wang2021real} benefits from GAN-based generation to restore more visually realistic images, some results still exhibit noticeable artifacts, as seen in the second and third results in the third row. Similarly, DiffTSR \cite{zhang2024diffusion} struggles to preserve realistic image qualities and often generates erroneous text structures.
In contrast, our method, with its strong SR prior and cross-attention mechanism, effectively restores text images with both high style realness and text fidelity. By generalizing well to real-world scenarios, our model produces more accurate and coherent results, surpassing the limitations of existing methods in handling real-world degradation.

{To provide additional qualitative evaluation, we conducted a human assessment study on a randomly selected subset of 100 images from our test dataset. The perceptual evaluation results demonstrate that our method generates images with acceptable quality in 70\% of cases, while DiffTSR achieves this standard in only 25\% of instances. This improvement in perceptual quality acceptance rate highlights the superior performance of our approach in generating visually plausible results.}

\subsection{Ablation Study}
In this subsection, we conduct an ablation study to evaluate the impact of different components in our proposed method.

\begin{wrapfigure}{r}{0.55\textwidth}
	\centering
	\includegraphics[width=\linewidth]{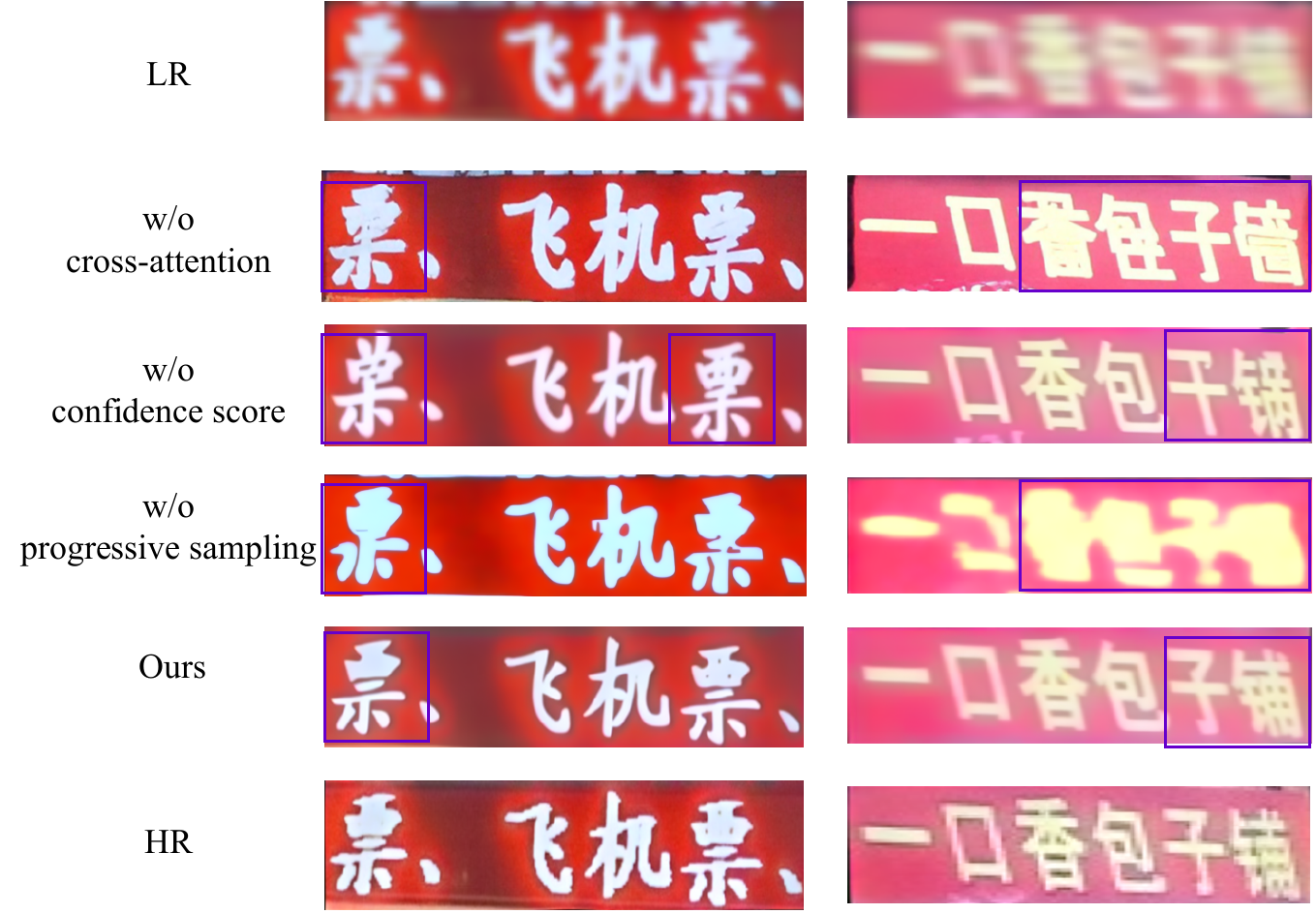}
	\caption{Ablation studies to validate the effectiveness of cross attention part, the confidence score mechanism and the data sampling strategy.}
	\label{fig:ablation}
	\vspace{-1em} 
\end{wrapfigure}

\noindent\textbf{Remove Cross-attention Module.} 
For the `w/o cross-attention' experiment where we remove the cross-attention module and perform image recovery using only image features.
As illustrated in the second row of Figure \ref{fig:ablation}, the characters are not well-formed and do not adhere to the correct structure of Chinese characters, leading to visually unpleasant results. This is because the model does not incorporate textual priors during the diffusion process, resulting in distorted character shapes and a lack of fidelity in the text.

\noindent\textbf{Remove Confidence Score.}
In the `w/o confidence score' experiment where we omit the confidence score from the OCR model but still use textual features, the performance improves over `w/o cross-attention', resulting in more accurate generated characters that are better aligned with the correct shapes, as shown in the third row of Figure \ref{fig:ablation}. However, some errors still persist. For instance, in the left image, the characters are partially correct but not entirely precise, and in the right image, some entirely incorrect characters are generated. This indicates that without the confidence score mechanism, the model cannot effectively correct errors in the generated text, as it lacks the feedback mechanism needed to refine the output during training.

\noindent\textbf{Without Progressive Sampling.}
In the `w/o progressive sampling' experiment, where LR images are directly employed to train the model, convergence proves to be challenging, as illustrated in the fourth row of Figure \ref{fig:ablation}. Training exclusively with LR images may not sufficiently expose the model to the contrasts between input and ground truth, thereby hindering its ability to learn the intricate features essential for high-quality image restoration.

In our full method, `Ours', which includes both the cross-attention module and the confidence score mechanism, the results improve significantly than these ablation settings. As shown in the fourth row of Figure \ref{fig:ablation}, the generated images exhibit more realistic visual appearances and accurate text structures. The cross-attention mechanism allows the model to better incorporate textual information into the image restoration process, while the confidence score module adjusts the importance of each character during training. This combination results in fewer errors and better overall text fidelity, demonstrating the effectiveness of both components.

Also, the quantitative results {on CTR-TSR-Test} are displayed in Table \ref{tab:ablation}. We can see that by leverging these components, our method is able to better utilize textual features, leading to more realistic and accurate text restoration. {Although our model utilizes a lightweight Unet module, it is still able to achieve satisfactory results after training.} This demonstrates the importance of the three parts.

\begin{table}[t]
    \centering
    \caption{Ablation Study on CTR-TSR-Test with $\times 4$ super-resolution to validate the effectiveness of cross attention part, the confidence score mechanism and the data sampling strategy.}
    \resizebox{0.8\columnwidth}{!}
    {
    \begin{tabular}{l|ccccc}
    \hline
    \multicolumn{1}{c|}{\multirow{2}{*}{method}} & \multicolumn{5}{c}{x4}        \\ \cline{2-6} 
    \multicolumn{1}{c|}{}                        & PSNR$\uparrow$ & LPIPS$\downarrow$ & FID$\downarrow$ & ACC$\uparrow$ & NED$\uparrow$ \\ \hline
    w/o cross-attention                                       &   21.54   &   0.250    &  25.26   &  0.7105   &   0.6934  \\
    w/o confidence score                                            &  21.83    &    0.185   &   7.150  &  0.7047   &  0.7136   \\
    w/o progressive sampling                                             &  19.31    &  0.245     &   20.22  &   0.7415  &  0.7023   \\
    Ours                                             & \textbf{23.50}&     \textbf{0.128}  & \textbf{5.583}    &  \textbf{0.8511}   &  \textbf{0.8723}   \\ \hline
    \end{tabular}
    }
    
    \label{tab:ablation}
    
    \end{table}

\begin{figure*}[t]
    \centering
     \includegraphics[width=0.9\linewidth]{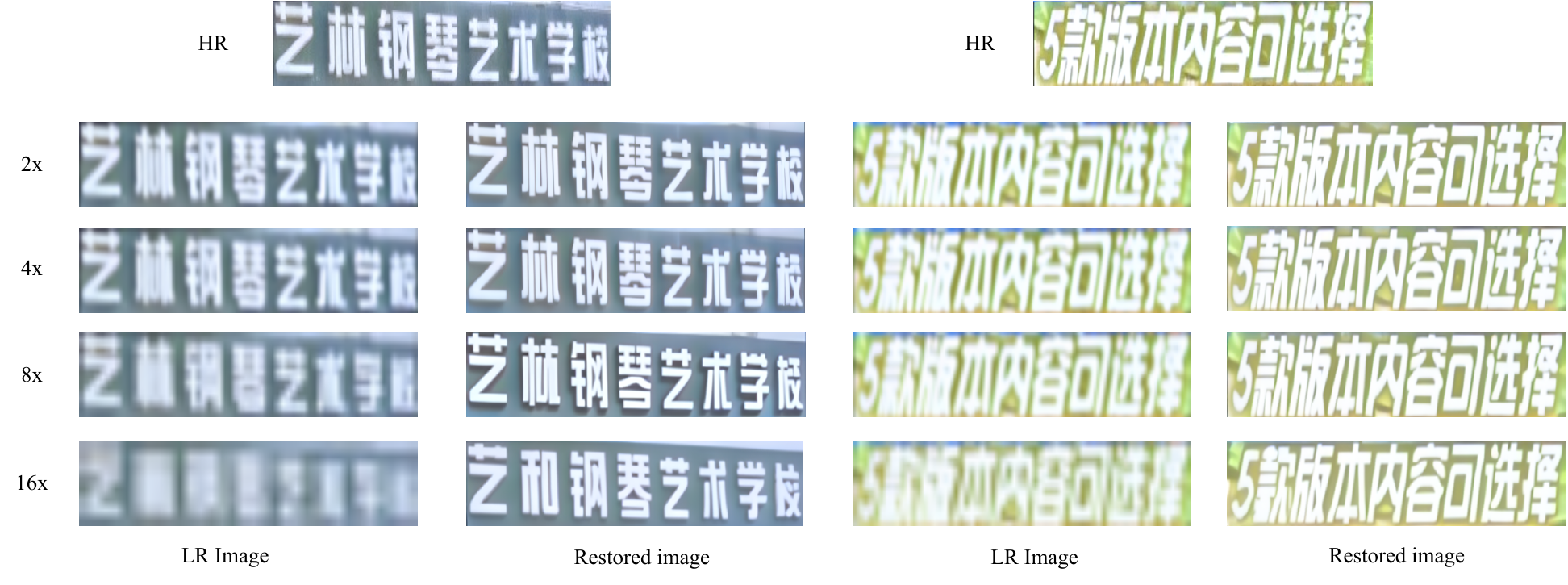}
     \vspace{-0.1in}
         \caption{Experiments to validate the robustness of our model towards dealing with various degrees of clarity.}
     \label{fig:sequence}
\end{figure*}

\subsection{Robustness of Our Model}
To assess the boundary performance of our model, we introduce a series of degradation operations followed by SR experiments on the degraded images. Specifically, the images are progressively downsampled and upsampled by factors of 2, 4, 8, and 16. The resulting image sequence is shown in Figure \ref{fig:sequence}. As demonstrated, when the scaling factor is below 16, our model excels in restoring the images, producing clear and high-quality results even under significant degradation. When the scaling factor reaches 16, the generated image still retains considerable quality, showing the robustness of our model in handling challenging degradation levels. These results underscore the model's effectiveness in recovering fine details and maintaining image clarity even at higher scaling factors, which sets it apart from traditional methods. This strong performance suggests that our model could be a promising solution for practical super-resolution tasks, where robustness is highly required.

%% file: sec/5_conclusion.tex
\section{Conclusion and Limitation}
In this paper, we introduce a novel approach to address the text image super-resolution problem by leveraging a strong super-resolution and reliable textual prior, which enhances the model's generalization capability. Our method begins with a data sampling strategy that gradually incorporates low-resolution images into the training process. 
Next, we integrate a powerful SR prior along with a cross-attention layer to incorporate textual features into the UNet architecture. To further improve performance, we employ a confidence score mechanism to minimize errors in the generated images. Extensive experimental results demonstrate that our model outperforms state-of-the-art methods, particularly in terms of style realness and text fidelity.

Despite the promising performance, our model still may produce errors, especially when handling small text in scenes with very low clarity. In future work, we plan to incorporate reinforcement learning techniques to refine and correct the generated text. Additionally, we aim to replace the OCR module with outputs from multi-modal large language models (MLLMs), which could potentially yield more accurate text outputs from the LR images.

%% file: sec/6_appendix.tex
\appendix
\onecolumn
\section{Training Details}
\vpara{Basic model information.} All the models are trained on the real-world dataset. LR and HR images are resized into $128\times 512\times 3$ and $32\times 128\times 3$, respectively. Besides, we utilize the TDM and MoM module from \cite{zhang2024diffusion} and load their pre-trained weights to process the text information and the text diffusion. When training the model, these two modules are frozen. For the Unet module, we utilize the pre-trained weights from the ResShift \cite{yue2024resshift} model. Also, to fuse the textual feature from the MoM module with the image feature, after each basic layers on the ResShift, we insert the Transformer layer for the cross attention module. The inserted Transformer layer is randomly initialized. During training, only the Unet module and the MoM module is updated. Besides, in the modeling of the text sequence, the maximum length of the text is set to 24 and all the characters in the text sequence belong to an alphabet with $K=6736$ characters including both Chinese and English characters as well as the numbers and special characters. The total step $T$ is set to 18.

\vpara{Data sampling strategy.} During training, the data sampling strategy is involved. Specifically, for the HR image $x_0$, in the first 10K training steps, we only use the degraded image $\hat{x_0}$ as the LR image, to build the model's core SR capabilities. In the next 10K training steps, we mix the LR image $y$ and $\hat{x_0}$ when constructing the HR-LR pairs to improve the ability of in-domain adaptation. Given the HR image $x_0$, the probabilities of using $y$ and $\hat{x_0}$ are equal. Finally, in the remaining training rounds, we extend the training process by introducing the degraded LR image $\hat{y}$ into training. We train the model for 400K steps in total.

\vpara{Deviations of Eq. (2).} According to Bayes's theorem, we have
\begin{equation}
    q(z_{t-1}|z_t, z_0, y)\propto q(z_t|z_{t-1},y)q(z_{t-1}|z_0,y),
\end{equation}
where
\begin{align}
    q(z_{t-1}|z_t,y)=\mathcal{N}(z_t;z_{t-1}+\alpha_te_0,\kappa^2\alpha_t\textbf{I}),\\
    q(z_{t-1}|z_0,y)=\mathcal{N}(z_{t-1};z_0+\eta_{t-1}e_0,\kappa^2\eta_{t-1}\textbf{I}).
\end{align}
We focus on the quadratic form in the exponent of $q(z_{t-1}|z_t,z_0,y_0)$, \emph{i.e.},
\begin{align}
    &-\frac{(z_t-z_{t-1}-\alpha_te_0)(z_t-z_{t-1}-\alpha_te_0)^T}{2\kappa^2\alpha_t}-\frac{(z_{t-1}-z_0-\eta_{t-1}e_0)(z_{t-1}-z_0-\eta_{t-1}e_0)^T}{2\kappa^2 \eta_{t-1}}\notag\\
    =&-\frac{1}{2}\left[\frac{1}{\kappa^2\alpha_t}+\frac{1}{\kappa^2\eta_{t-1}}\right]z_{t-1}z_{t-1}^T+\left[\frac{z_t-\alpha_te_0}{\kappa^2\alpha_t}+\frac{z_0+\eta_{t-1}e_0}{\kappa^2\eta_{t-1}}\right]z_{t-1}^T+\text{const}\notag\\
    =&-\frac{(z_{t-1}-\mu)(z_{t-1}-\mu)^T}{2\lambda^2}+\text{const},
\end{align}
where
\begin{equation}
    \mu=\frac{\eta_{t-1}}{\eta_t}z_t+\frac{\alpha_t}{\eta_t}z_0, \lambda^2=\kappa^2\frac{\eta_{t-1}}{\eta_t}\alpha_t,
\end{equation}
and const denotes the item that is independent of $z_{t-1}$. After substituting $z_0$ with the neural network $f_{\theta}(z_t, y, t)$, this quadratic form induces the Gaussian distribution of Eq. (6).

\section{Procedure of Our Framework}
The training and inference procedure of our framework can be summarized in Algorithm \ref{alg:1} and Algorithm \ref{alg:2}, respectively.

\begin{algorithm}[h] 
    \caption{Training of Our Framework}
    \begin{algorithmic}[1]
        \State \text{Initialize Training step} $r=0$
        \Repeat
            \State $x_0,y,c_0\sim q(x_0,y,c_0)$, where $c_0$ is the ground truth text on $x_0$.
            \State $z_0=E(x_0)$, where $E$ is the VQGAN encoder.
            \State Generate $\hat{x_0}$ and $\hat{y}$ via the pipeline in ESRGAN \cite{wang2021real}.
            \If{$r < 10000$}
                \State $y'=\hat{x_0}$
            \ElsIf{$10000\leq r<20000$}
                \State $y'=\text{Uniform}(\hat{x_0}, y)$
            \Else
                \State $y'=\text{Uniform}(\hat{x_0}, y, \hat{y})$
            \EndIf
            \State $\vec{\alpha}, c=\text{OCR}(y')$
            \State $z_y=E(y')$
            \State $\epsilon\sim \mathcal{N}(0,I)$
            \State $t\sim \text{Uniform}(\{1,2,\cdots,T\})$
            \State $z_t=\sqrt{\bar{\alpha_t}}z_0+\sqrt{1-\bar{\alpha_t}}\epsilon$
            \State $c_t\sim \mathcal{C}(c_t|\bar{\alpha_t}c_0+\frac{1-\bar{\alpha_t}}{K})$
            \State $[\textbf{I}_{cond_t}, \textbf{C}_{cond_t}]=\text{MoM}([z_y, z_t], \vec{\alpha}\cdot c_t, t)$
            \State $\bar{z_0}=f_{\theta}(z_t, y, t, \textbf{C}_{cond_t})$
            \State $\bar{x_0}=D(\bar{z_0})$, where $D$ is the VQGAN decoder.
            \State $\bar(c_0)=\text{OCR}(\bar{x_0})$
            \State $\mathcal{L}=\lambda_1\mathcal{L}_{L1}(x_0,\bar(x_0))+\lambda_2\mathcal{L}_{LPIPS}(x_0,\bar(x_0))+\lambda_3\mathcal{L}_{CE}(c_0,\bar(c_0))$
            \State Take gradient descent on $\nabla_{\theta} \mathcal{L}$
        \Until{Converged}
    \end{algorithmic}
    \label{alg:1}
\end{algorithm}

\begin{algorithm}
    \caption{Inference of our Framework}
    \begin{algorithmic}[1]
        \Require LR image $y$
        \Ensure HR image $x$
        \State $z_y = E(y)$
        \State $\epsilon\sim\mathcal{N}(0,I)$
        \State $z_T=z_y+\kappa \epsilon$
        \State $c_T=\text{OCR}(y)$
        \For{$t=T\cdots 1$}
            \State $z\sim \mathcal{N}(0,I)$ if $t>1$, else $z=0$
            \State $[\textbf{I}_{cond_t}, \textbf{C}_{cond_t}]=\text{MoM}([z_y, z_t], \vec{\alpha}\cdot c_t, t)$
            \State Sample $z_{t-1}$ via Eq. (6).
            \State $c_{pred,t}=\tau(c_t, \textbf{I}_{cond_t}, t)$, where $\tau$ is the transformer decoder intended for the text diffusion module.
            \State $\tilde{\pi}=\left[\alpha_t c_t+\frac{1-\alpha_t}{K}\right]\odot\left[\bar{\alpha}_{t-1}+\frac{1-\bar{\alpha}_{t-1}}{K}\right]$
            \State $\pi_{post}(c_t, c_{pred, t})=\frac{\pi}{\sum_{k=1}^{K}\pi_k}$
            \State $c_{t-1}\sim \mathcal{C}(c_{t-1}|\pi_{post}(c_t,c_{pred,t}))$ if $t>1$ else $c_{0}\sim \mathcal{C}(c_0|c_{pred, t})$
        \EndFor
        \State \Return $x=D(z_0)$ where $D$ is the VQGAN decoder.
    \end{algorithmic} \label{alg:2}
\end{algorithm}

\section{Efficiency of Our model}
To demonstrate the efficiency of our model, we present the total number of parameters and the average inference time per image in Table \ref{tab:efficiency}. For the average inference time, we randomly select 100 LR text images from the test set with the size of $128\times 512$ as the input. Compared with the state-of-the-art diffusion-based text image SR method, DiffTSR, our model incorporates a lightweight SR prior, leading to an 82.2\% reduction in parameters. Furthermore, our model achieves a 93.4\% reduction in inference time, significantly improving efficiency over DiffTSR. These results highlight the effectiveness of our approach in balancing performance and computational cost.

\begin{table}[h]    
    \centering
    \caption{the total number of parameters and the average inference time on 100 randomly selected LR images of size $128\times 512$ for both our model and DiffTSR.}
    \begin{tabular}{c|c|c}
    \hline
    \multicolumn{1}{l|}{} & \# parameters & Inference Time(s)                     \\ \hline
    DiffTSR               & 874M &         25.32             \\ \hline
    Ours                  & 155M & 1.68 \\ \hline
    \end{tabular}
    
    \label{tab:efficiency}
\end{table}

\section{Future Works}
Due to the limited performance of the OCR module, existing methods still make mistakes in the generated images. Therefore, in future work, we plan to incorporate reinforcement learning techniques to refine and correct the generated text, such as DPO \cite{black2023training}. Additionally, we aim to replace the OCR module with outputs from multi-modal large language models (MLLMs) \cite{liu2023visual,lu2024deepseek}, which could potentially yield more accurate text outputs from the LR images. 
\section{More Visual Results}
\begin{figure*}[h]
    \centering
     \includegraphics[width=1\linewidth]{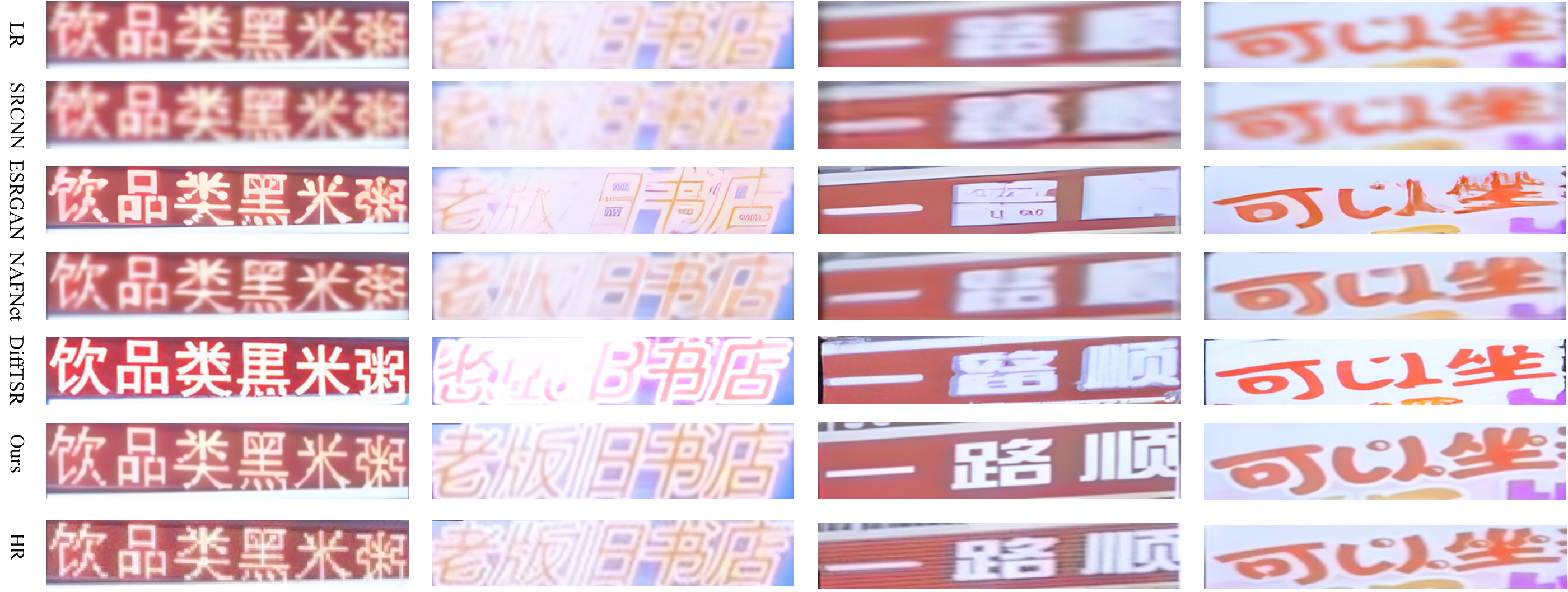}
         \caption{Qualitative comparison on the real-world dataset with different methods including SRCNN \cite{dong2015image}, NAFNet \cite{chen2022simple}, ESRGAN \cite{wang2021real}, the existing SOTA method DiffTSR \cite{zhang2024diffusion} and our method for $\times 4$ super-resolution. }
     \label{fig:tifo3}
\end{figure*}

\begin{figure*}[h]
    \centering
     \includegraphics[width=1\linewidth]{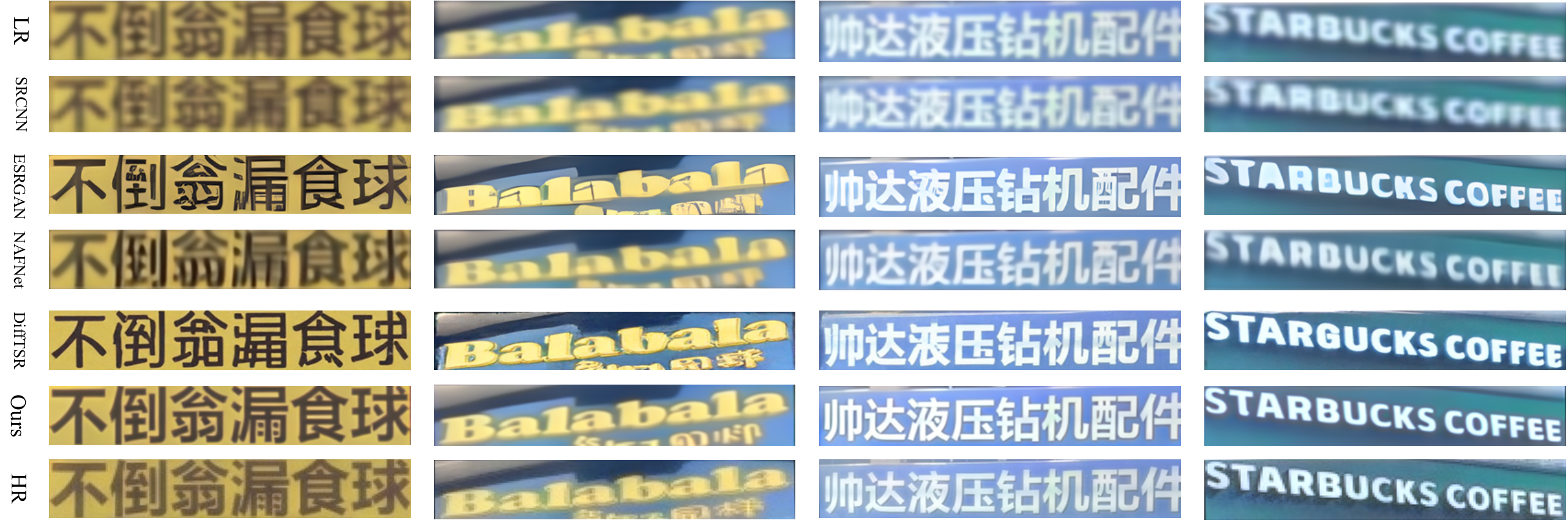}
         \caption{Qualitative comparison on the real-world dataset with different methods including SRCNN \cite{dong2015image}, NAFNet \cite{chen2022simple}, ESRGAN \cite{wang2021real}, the existing SOTA method DiffTSR \cite{zhang2024diffusion} and our method for $\times 4$ super-resolution. }
     \label{fig:tifo4}
\end{figure*}

\begin{figure*}[h]
    \centering
     \includegraphics[width=1\linewidth]{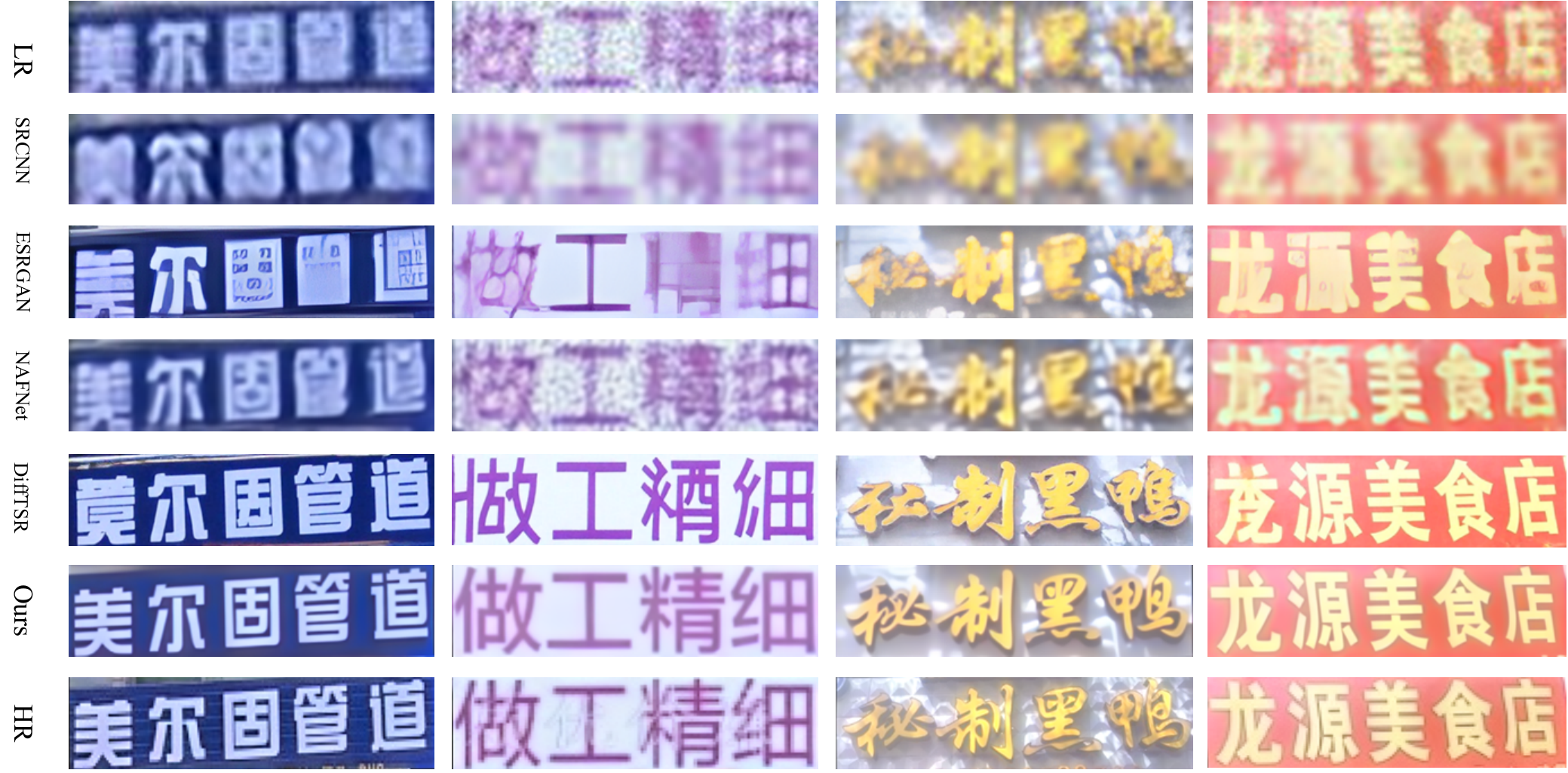}
         \caption{Qualitative comparison on the synthetic dataset with different methods including SRCNN \cite{dong2015image}, NAFNet \cite{chen2022simple}, ESRGAN \cite{wang2021real}, the existing SOTA method DiffTSR \cite{zhang2024diffusion} and our method for $\times 4$ super-resolution. }
     \label{fig:tifo5}
\end{figure*}

More visual results are shown in this section for the real-world datasets in Figure \ref{fig:tifo3} and Figure \ref{fig:tifo4}. The visual results on the synthetic dataset are shown in Figure \ref{fig:tifo5}. Visual results show that our method effectively handle the LR images from the real-world scenarios that contains English letters, numbers and diverse text styles.

